# Egyptian Sign Language Recognition Using CNN and LSTM

Ahmed Adel Gomaa Elhagry, Rawan Gla Elrayes
Comp. & Sys. Dept., Faculty of Engineering, Zagazig University, Egypt,
ahmedelhagary99@gmail.com
rewan25299@gmail.com
Supervisors: Dr.Amr Zamel, Comp. & Sys. Dept, Faculty of Engineering, Zagazig University, Egypt, *eng.amrezml@gmail.com*.
Dr.Ahmed Helmy, Comp. & Sys. Dept, Faculty of Engineering, Zagazig University, Egypt

## I. ABSTRACT

Sign language is a set of gestures that deaf people use to communicate. Unfortunately, normal people don't understand it, which creates a communication gap that needs to be filled.
Because of the variations in (Egyptian Sign Language) ESL from one region to another, ESL provides a challenging research problem.
In this work, we are providing applied research with its video-based Egyptian sign language recognition system that serves the local community of deaf people in Egypt, with a moderate and reasonable accuracy. We present a computer vision system with two different neural networks architectures. The first is a Convolutional Neural Network (CNN) for extracting spatial features. The CNN model was retrained on the inception mod. The second architecture is a CNN followed by a Long Short Term Memory (LSTM) for extracting both **spatial and temporal features**. The two models achieved an accuracy of 90% and 72%, respectively. We examined the power of these two architectures to distinguish between 9 common words (with similar signs) among some deaf people community in Egypt.

## II. INTRODUCTION

In 2019, a study showed that 466 million people worldwide have disabling hearing loss [1].
In Egypt, the percentage reached about 7% of our population. More than 7 million Egyptians are being marginalized; special elementary schools with a poor quality of education and a lack of supervision, universities that don't accept them, and job vacancies that refuses them just because they are deaf.

In addition, most of their parents tend to keep them indoors and avoid socializing. Also, Whenever they start using sign language in public places, people either stare at them, or talk normally, ignoring the fact that they actually need to communicate and help.

Sign language is their only tool to express what they really need, and without an interpreter, communication between deaf and others remains a barrier that prevents them from having a normal life.
So what is sign language? Sign language is a set of hand gestures, facial expressions and body motion that represent words. Each country has its own sign language. In Egypt, ESL varies from one region to another with a percentage of about 20%. Up to date, there is no official documented, updated resource for ESL vocabulary. Also, ESL is continuously updating and available online datasets aren't used by the deaf community. This adds a notable difficulty for any trial to use modern technology in developing a suitable and efficient tool to help this respectable community of deaf people. On the other hand, facilitating communication between deaf people and normal individuals aims to eliminate the need for a deaf person to have a human interpreter everywhere, developing the work in deaf schools and emerging them into common work tasks like normal people.

This work is focused on providing a real time communication system with an updated ESL dataset to serve deaf people in Egypt. Our video dataset has been collected in Zagazig, Egypt.
First, each video is subsampled to frames, then the features are extracted from the frames, finally, each frame is classified into one of the 9 words in our dataset using neural networks.

This work is based on the paper: Real-Time Sign Language Gesture (Word) Recognition from Video Sequences Using CNN and RNN[2]

This paper is organized as follows: Section 2 provides a quick view of related work. Section 3 introduces the proposed technique where both tools and dataset are well explained. Section 4 gives a summary of obtained experimental results. Finally, conclusions and future work are given in Section 5.

## II. RELATED WORK

Research in sign language recognition has gained a notable interest since many years ago. Nowadays, modern technologies in handheld and smart devices facilitates a lot many processes in computer vision tasks. Also, various programming languages become rich of off-the-shelf packages and source codes, in particular those of developing mobile apps. Most researchers have been following one of three approaches: sensor-based gloves, 3-D skeletons, or computer vision.

The first two approaches neglect facial expressions which play a huge rule in sign language recognition. On the other hand, computer-vision systems are capable of capturing the whole gesture, not to mention their mobility that differentiates them from glove-based systems.

Reference [3] proposed a computer vision approach for continuous American sign language recognition (ASL). He used a single camera to extract two-dimensional features as input of the Hidden Markov Model (HMM) on a dataset of 40 words collected in a lab. He followed two approaches for the camera position: desk-mounted cam with a word recognition accuracy of 92% and wearable cap-mounted cam with an accuracy of 98%. Another computer vision system developed by Dreuw et al. was able to recognize sentences of continuous sign language independent of the speaker, described in Reference [4]. He employed pronunciation and language models in sign language with a recognition algorithm based on the Bayes' decision rule. The system was tested on a publicly available benchmark database consisting of 201 sentences and 3 signers, and they achieved a 17% word error rate.

Reference [5] also proposed a computer vision system that uses hand and face detection for classifying ASL alphabets into four groups depending on the hand's position. The system used the inner circle method and achieved an accuracy of 81.3%.

Reference [6] followed a different approach for classifying ASL alphabets. They used the Leap

Motion controller to compare the performance of the Naive Bayes Classifier (NBC) with a Multilayer Perceptron (MLP) trained by the backpropagation algorithm. An accuracy of about 98.3% was achieved using NBC, while MLP gave an accuracy of about 99.1%

In the past 20 years, deep learning have been used in sign language recognition by researchers from all around the world.
Convolutional Neural Networks (CNNs) have been used for video recognition and achieved high accuracies last years. B. Garcia and S. Viesca at Stanford University proposed a real-time ASL recognition with CNNs for classifying ASL letters, described in Reference
[7] . After collecting the data with a native camera on a laptop and pre-training the model on GoogLeNet architecture, they attained a validation accuracy of nearly 98% with five letters and 74% with ten.
CNNs have also been used with Microsoft Kinect to capture depth features. Reference [8] proposed a predictive model for recognizing 20 Italian gestures. After recording a diverse video dataset, performed by 27 users with variations in surroundings, clothing, lighting and gesture movement, the model was able to generalize on different environments not occurring during training with a cross-validation accuracy of 91.7%.
The Kinect allows capture of depth features, which aids significantly in classifying ASL signs. Reference [9] also used a CNN model with kinect for recognizing a set of 50 different signs in the Flemish Sign Language with an error of 2.5%. however, this work considered only a single person in a fixed environment.

There's another version of CNNs called 3D CNNs, which was used by Reference [10] to recognize 25 gestures from Arabic sign language dictionary. The recognition system was fed with data from depth maps. The system achieved 98% accuracy for observed data and 85% average accuracy for new data.

Computer vision systems face two major challenges: environmental concerns (e.g. lighting sensitivity, background) and camera's position. Most previous systems lack the diversity of data and capturing the whole gesture.

### III. PROPOSED METHOD

In this work, we propose a computer-vision system for the task of sign language recognition.

Our proposed method doesn't depend on using glove-based sensors because hand gestures are only a part of sign language. Instead, it captures the hand, face, and body motion. In addition, unlike most previous research, our dataset was collected in various real backgrounds and lightning conditions rather than the lab.

Deep learning can be divided into 2 categories: supervised and unsupervised.
In supervised learning, the data is labeled during training. Choosing the network type and architecture depends on the task at hand.
CNN's have been used for isolated image recognition. However, for continuous recognition, other architectures such as CNN-LSTM are more convenient.

Transfer learning is a technique that shortcuts much of this by taking a piece of a model that has already been trained on a related task and reusing it in a new model.
Modern image recognition models have millions of parameters. Training them from scratch requires a lot of labeled training data and a lot of computing power (hundreds of GPU-hours or more).
In this work, We have used Inception-v3 architecture of GoogLeNet[11] to extract spatial features from the frames of video sequences to classify a set of 9 Egyptian Sign Language gestures.

The pre-trained Inception-v3 is a widely-used image recognition deep learning model with millions of parameters. The model achieved state-of-the-art accuracy for recognizing general objects with 1000 classes in the ImageNet dataset[12]. . The model extracts general features from input images in the first part and classifies them based on those features in the second part. It is based on the original paper[13]. Fig 1 Shows a visualization.

I'll propose two different network architectures to classify video-gesture-data into words: CNNs and LSTMs.

After extracting frames from each video, we pass the features vector as an input to the first layer of the neural network.

One of the major challenges we faced was the lack of real data, or at least some trusted datasets.

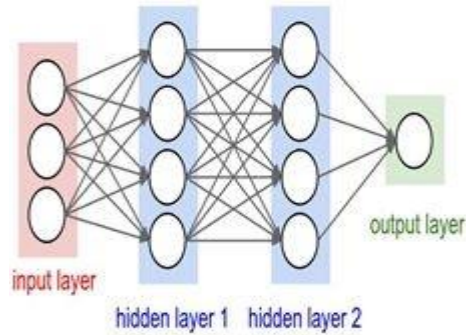

Fig 2

Each neuron receives some inputs, performs a dot product and optionally follows it with a non-linearity. The whole network still expresses a single differentiable score function: from the raw image pixels on one end to class scores at the other. In the last layer (fully-connected), The number of neurons equals the number of classes, and it has a loss function (e.g. SVM/Softmax) that determines how good the algorithm is doing ( See Fig 2).

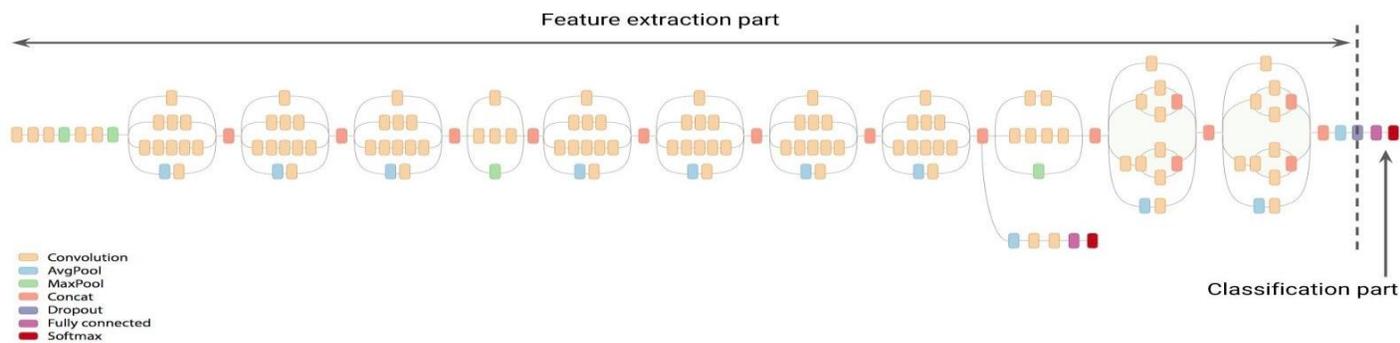

Fig1: Inception v3

**Convolutional Neural Networks (CNNs).**
CNNs are one of the most popular methods in classifying images.
CNNs are made up of layers of neurons that have learnable weights and biases. Adjusting these weights properly, the layers are able to extract features from the input image. The CNN takes an image (a vector representation of an

**Long-Short-Term memory (LSTM).**

Convolutional neural networks are great for a 1 to 1 relation; given an image of a sign, it generates fixed-size label, like the class of the sign in the image. However, What CNNs cannot do is accept a sequence of vectors. That's where Recurrent Neural Networks (RNNs) are used. RNNs allow us to understand the context of a video frame, relative to the frames that came before it. They do this by passing the output of one training step to the input of the next training step, along with the new frame.

We're using a special type of RNN here, called an LSTM, that allows our network to learn long-term dependencies. Figure 2 shows a visualization for an LSTM cell.

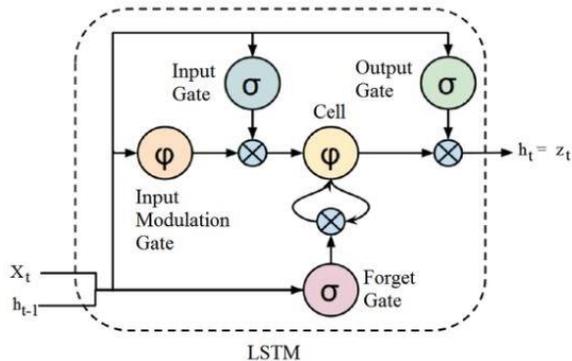

Fig 3- Basic LSTM cell

The CNN-LSTM architecture involves using a pre-trained CNN for feature extraction from input data along with LSTMs for sequence prediction

It is helpful to think of this architecture as defining two sub-models: the CNN Model for feature extraction and the LSTM Model for interpreting the features across time steps.
We used to methods;
The first method we followed is pre-training the inception v3 model on our data.

The second method is to pass the predicted labels from the Inception CNN model to an LSTM. The last hidden layer of the CNN is the input to the LSTM.

After extracting the bottleneck features, we used a network consisting of single layer of 512 LSTM units, followed by a fully connected layer with Softmax activation. Finally, a regression layer is applied. For minimizing the loss function, We used Adaptive Moment Estimation (ADAM) [14] which is a stochastic optimization algorithm that's been designed for training deep neural networks, and we trained the data for 10 epochs.

## IV. EXPERIMENTS and RESULTS

As mentioned before, there isn't an official, updated ESL resource. So we created our own updated dataset by visiting "Hope School For The Deaf" in Zagazig-Egypt, and collecting 9 words from a volunteer deaf student: "similar, differ, doctor, free, I love you, judge, Sunday, Talk, Travel".
"متشابه, مختلف, طبيب, فاضي, أحبك, قاضي, الأحد", يتكلم, مسافر."

**We have chosen these words because their signs are similar and we wanted to test the performance of the algorithm on distinguishing between them.*show them***
Afterward, the dataset was prepared by capturing videos in various lighting conditions, angles, and camera positions.
Our dataset has 9 classes, 2 signers, 5 places, 20 videos per place, 100 videos per class, thus we get 900 videos in total.
Using CNN only, we achieved 90% accuracy, while passing the predicted labels from the Softmax layer of our CNN to our LSTM gave an accuracy of 72%.

## V. CONCLUSIONS

In this work, we have presented a vision-based system to translate some Egyptian Sign Language gestures to their alternative isolated words. We used two different architectures for classification; Inception v3 CNN and Inception v3 CNN-LSTM.

We obtained an accuracy of 90% using the first architecture. While the Inception v3 CNN-LSTM architecture resulted in an accuracy of 72%. This shows that CNN give high accuracies for isolated sign language recognition, while CN-LSTM is a great choice for continuous word recognition.

## VI.  FUTURE WORK

It will be interesting to continue what we have started in this work by collecting more signs from ESL to widen our dataset of sentences for the process of continuous classification.

Also, it is reasonable to try other GoogleNet models like VGG16 and AlexNet.

Our most important goal is to implement one mobile app around our findings in this field where it will highly facilitate the communication between deaf people and others everywhere. The preliminary ideation is to capture a video for one ELS using the mobile camera then processing it online in a backend software which implements our model. It seems challenging as more ELS signs added to the database but the great help and valuable gain of such application are so worthy.

## V.  ACKNOWLEDGMENT

Many thanks to **"El-Amal School For The Deaf"** in Zagazig-Egypt for their great help in collecting sign language data . We are thankful to both Dr. Amr Zamel and Dr. Ahmed Helmi for their help and encouragement.